\documentclass[sigconf,screen]{acmart}

\usepackage{graphicx}

\usepackage{algorithm}
\usepackage{algpseudocode} 
\usepackage[most]{tcolorbox}
\usepackage{diagbox}
\usepackage{multirow}
\usepackage{stmaryrd}
\usepackage{subcaption}

\usepackage{array}
\usepackage{booktabs}
\usepackage{tikz}
\usetikzlibrary{calc, positioning}
\usepackage{todonotes}
\usepackage{breqn}
\usepackage{todonotes}
\usepackage{amsthm}
\usepackage[utf8]{inputenc} 
\usepackage[T1]{fontenc}    
\usepackage{hyperref}       
\usepackage{url}            
\usepackage{booktabs}       
\usepackage{nicefrac}       
\usepackage{microtype}      
\usepackage{xcolor}         
\usepackage{listings} 
\usepackage{multirow}
\usepackage{makecell} 
\lstdefinestyle{custom}{
    backgroundcolor=\color{gray!3}, 
    basicstyle=\ttfamily\small,     
    keywordstyle=\color{blue}\bfseries, 
    commentstyle=\color{green!60!black}, 
    stringstyle=\color{red!70!black}, 
    frame=single,                   
    numbers=left,                   
    numberstyle=\tiny\color{gray},  
    breaklines=true,                
    captionpos=b,                   
    frame=none
}

\definecolor{codegray}{gray}{0.95}
\definecolor{commentgreen}{rgb}{0,0.5,0}
\definecolor{keywordblue}{rgb}{0.0,0.0,0.5}

\lstdefinelanguage{LawSpec}{
    morekeywords={G, F, U, ~, &, |, ->},
    sensitive=true,
    morecomment=[l]{//},
    morestring=[b]",
}

\lstdefinestyle{lawstyle}{
    language=LawSpec,
    backgroundcolor=\color{codegray},
    basicstyle=\ttfamily\small,
    keywordstyle=\color{keywordblue}\bfseries,
    commentstyle=\color{commentgreen}\itshape,
    breaklines=true,
    showstringspaces=false,
    columns=fullflexible
}

\newcommand{\tool}{\textsc{ProbGuard}}
\newcommand{\rev}[1]{ #1}

\acmConference[ASE '26]{Proceedings of the 41st IEEE/ACM International Conference on Automated Software Engineering}{October 12--16, 2026}{Munich, Germany}
\acmBooktitle{Proceedings of the 41st IEEE/ACM International Conference on Automated Software Engineering (ASE '26), October 12--16, 2026, Munich, Germany}

\begin{document}

\title[\tool: Proactive Runtime Monitoring for LLM Agent Safety via Probabilistic Prediction]{\tool: \rev{Proactive} Runtime Monitoring for LLM Agent Safety via Probabilistic Prediction}

\author{Haoyu Wang}
\orcid{0009-0000-6379-5312}
 \affiliation{%
   \institution{Singapore Management University}
   \country{Singapore}
}
\email{haoyu.wang.2024@phdcs.smu.edu.sg}

\author{Christopher M. Poskitt}
\orcid{0000-0002-9376-2471}
\affiliation{\institution{Singapore Management University}\country{Singapore}}
\email{cposkitt@smu.edu.sg}

 \author{Jiali Wei}
 \orcid{0000-0003-1842-0196}
 \affiliation{%
   \institution{Xi'an Jiaotong University}
   \country{Xi'an, China}
   }
 \email{weijiali1119@stu.xjtu.edu.cn}

\author{Jun Sun}
\orcid{0000-0002-3545-1392}
\affiliation{\institution{Singapore Management University}\country{Singapore}}
\email{junsun@smu.edu.sg}

\begin{abstract}
Large Language Model (LLM) agents increasingly operate across domains such as robotics, virtual assistants, and web automation. 
However, their stochastic decision-making introduces safety risks that are difficult to anticipate during execution. 
Existing runtime monitoring frameworks, such as AgentSpec, primarily rely on reactive safety rules that detect violations only when unsafe behavior is imminent or has already occurred, limiting their ability to handle long-horizon dependencies.
We present \tool{}, a proactive runtime monitoring framework for LLM agents that anticipates safety violations through probabilistic risk prediction. 
\tool{} abstracts agent executions into symbolic states and learns a Discrete-Time Markov Chain (DTMC) from execution traces to model behavioral dynamics. 
At runtime, the monitor estimates the probability that execution will remain safe from the current state, and triggers an intervention when this probability falls below a user-defined threshold. 
To improve robustness, \tool{} incorporates semantic validity constraints in the abstraction and admits a PAC-style analysis that characterizes the sample complexity required to certify the learned model under standard assumptions.
We evaluate \tool{} in two safety-critical domains: autonomous driving and embodied household agents. 
Across evaluated scenarios, \tool{} consistently predicts traffic law violations and collisions in advance, with warnings up to 15.84 seconds at a threshold yielding no false alarms, and up to 38.66 seconds at stricter thresholds. 
In embodied agent tasks, \tool{}'s re-prompting intervention mode reduces unsafe behavior by 65.37\% relative to the unmonitored baseline while retaining 80.4\% of the baseline task completion; a stricter halting configuration reduces unsafe behavior by 93.60\% at a larger cost in completion.
\tool{} is implemented as an extensible open-source runtime monitor integrated with the LangChain agent framework and introduces minimal runtime overhead.
\end{abstract}



\maketitle


\section{Introduction}

Large Language Models (LLMs) are increasingly used as the foundation for building autonomous agents.
These agents function as programmable entities operating across domains ranging from code generation and productivity tools to embodied household tasks and robotics~\cite{agentbench,shinn2023reflexion,lu2024codeact}.
Unlike traditional software systems, which are designed around explicit specifications and fixed control logic, LLM-powered agents interpret natural language goals, synthesize action plans, and adaptively respond to environmental feedback.
In this paradigm, the agent itself effectively becomes the software~\cite{shinn2023reflexion}.

However, this autonomy introduces serious safety concerns~\cite{wang2025agentspec}.
LLM agents may take harmful actions, misinterpret ambiguous instructions, or behave inconsistently under minor context shifts~\cite{ribeiro2020beyond,lin2023truthfulqa,ouyang2022training}. 
These issues resemble familiar software engineering challenges such as bugs, specification gaps, and nondeterminism, but are amplified by the opacity and adaptivity of foundation models.
In high-stakes domains such as cyber-physical systems, sensitive data management, or decision pipelines, such risks become particularly critical~\cite{liang2022holistic,agentbench}.
Failures may manifest in subtle yet consequential ways, for example skipping confirmation steps in safety-critical workflows, misclassifying objects before manipulation, or granting elevated privileges due to ambiguous instructions.
Consequently, ensuring the trustworthy deployment of LLM agents requires systematic runtime monitoring mechanisms capable of detecting and mitigating unsafe behavior during execution~\cite{zhang2024fusionlargelanguagemodels,zhang2026llmenabledapplicationsrequiresystemlevel}.

To improve agent reliability, several frameworks introduce oversight layers that monitor agent behavior during execution.
For example, AgentSpec~\cite{wang2025agentspec} and GuardAgent~\cite{xiang2025guardagentsafeguardllmagents} implement rule-based monitors that track agent actions against interpretable safety constraints, such as preventing unauthorized access to sensitive records.
To account for contextual uncertainty, ShieldAgent~\cite{chen2025shieldagentshieldingagentsverifiable} incorporates Markov logic networks, allowing the monitor to evaluate rule relevance probabilistically.
However, these approaches are largely \emph{reactive}: violations are detected only when a specific state transition breaches a safety rule or when a violation becomes imminent.
Such reactive monitoring lacks the temporal foresight needed to manage long-horizon risks.
For instance, in autonomous driving, a rule stating that a vehicle must not collide with another vehicle provides little opportunity for intervention once the collision becomes unavoidable.
A more effective approach is to detect risk earlier, when the system can anticipate that the agent is following an unsafe trajectory, such as accelerating toward a busy intersection without sufficient braking distance.

To address this limitation, we propose \tool{}, a framework that shifts agent oversight from reactive monitoring to proactive runtime monitoring through probabilistic risk prediction.
\tool{} operates through a multi-stage monitoring pipeline (Figure~\ref{fig:pro2guard-workflow}).
Offline, it collects execution traces, applies domain-specific predicate abstraction~\cite{10.5555/647766.733618} to map trajectories into symbolic states, and learns a Discrete-Time Markov Chain (DTMC) over those states, enabling reasoning about future behavior over long horizons; Probably Approximately Correct (PAC) analysis~\cite{bazille2020global} characterizes how far the learned dynamics may deviate from the true ones under finite sampling.
At runtime, the monitor estimates the probability that execution will remain safe from the current state and intervenes when this probability falls below a user-defined threshold, adapting agent behavior before unsafe states are reached.
\rev{\tool{} is designed as a policy-agnostic runtime monitor: the framework makes no assumption about how the monitored agent selects actions. The underlying policy may be LLM-driven, neural network-based, rule-based, or hybrid. \tool{} assumes only the observation of the agent's behavior in the form of a sequence of concrete states (that are subsequently processed through predicate-based abstraction).}

\begin{figure}[t]
\centering
\footnotesize
\resizebox{\linewidth}{!}{
\begin{tikzpicture}[
    node distance=1.2cm and 0.8cm,
    box/.style={
        draw,
        rounded corners,
        align=center,
        minimum width=2.2cm,
        minimum height=0.95cm
    },
    sbox/.style={
        draw,
        rounded corners,
        align=center,
        minimum width=2.0cm,
        minimum height=0.9cm
    },
    line/.style={->, thick},
    every node/.style={font=\footnotesize}
]


\node[box, fill=gray!10] (traces) {Execution\\traces};
\node[box, fill=gray!10, right=of traces] (abstraction) {Predicate\\abstraction};
\node[box, fill=gray!10, right=of abstraction] (dtmc) {Learned\\DTMC};
\node[box, fill=gray!10, right=of dtmc] (spec) {Safety\\property};

\draw[line] (traces) -- (abstraction);
\draw[line] (abstraction) -- (dtmc);
\draw[line] (spec.west) -- (dtmc.east);


\node[box, fill=blue!8, below=1.8cm of abstraction, xshift=-2.0cm] (agent)
{LLM agent\\+ environment};

\node[sbox, fill=blue!8, right=of agent] (state)
{Symbolic\\state};

\node[sbox, fill=blue!8, right=of state] (risk)
{Risk prediction\\$P_{\mathrm{safe}}$};

\node[sbox, fill=blue!8, below=1.1cm of risk] (alert)
{Risk alert /\\intervention};

\draw[line] (agent) -- node[above]{observe} (state);
\draw[line] (state) -- (risk);

\draw[line] (risk) -- node[right]{\small $<\theta$} (alert);

\draw[line] (alert.west) -| node[pos=0.25, below]{} (agent.south);


\coordinate (join) at ($(risk.north)+(0,0.8)$);

\draw[-, thick, dashed] (dtmc.south) |- (join);
\draw[-, thick, dashed] (spec.south) |- (join);
\draw[->, thick, dashed] (join) -- (risk.north);


\node[align=center, font=\bfseries]
at ($(traces.north)!0.5!(spec.north)+(0,0.8)$)
{Offline model construction};

\node[align=center, font=\bfseries]
at ($(agent.south)!0.5!(risk.south)+(0,-0.9)$)
{Online runtime monitoring};

\end{tikzpicture}
}
\caption{High-level workflow of \tool{}. Offline, the framework learns a probabilistic model from execution traces and domain-specific abstractions. Online, it abstracts the current agent state, estimates the probability that execution will remain safe, and issues an alert or intervention when that probability falls below the threshold~$\theta$.}
\label{fig:pro2guard-workflow}
\end{figure}
 
We evaluate \tool{} in two safety-critical domains characterized by stochastic environments: autonomous vehicles and embodied household agents. 
To ground the evaluation, we derive domain-specific safety properties from LawBreaker~\cite{sun2022lawbreaker} (traffic rules for autonomous driving) and SafeAgentBench~\cite{yin2025safeagentbenchbenchmarksafetask} (object- and state-level safety rules for embodied agents), from which we extract predicates that define the abstract state space.
In autonomous driving scenarios, \tool{} integrates a monitor automaton synchronized with the learned DTMC to predict potential violations in future driving trajectories.
Across evaluated scenarios, \tool{} consistently predicts traffic law violations and collision risks in advance, providing warnings up to 15.84 seconds before violations occur at a threshold that raises no false alarms, and up to 38.66 seconds under stricter thresholds.
In embodied agent tasks, the re-prompting intervention mode reduces unsafe behavior by 65.37\% relative to the unmonitored baseline while preserving 80.4\% of the baseline task completion; the strictest halting configuration reduces unsafe behavior by 93.60\% at a larger cost in completion.
Moreover, \tool{} introduces minimal runtime overhead, staying below 50\,ms for embodied agents and approximately 100\,ms for autonomous driving scenarios.

The contributions of this work are summarized as follows:
\begin{itemize}
    \item \textbf{Proactive probabilistic runtime monitoring for LLM agents.}
    We present \tool{}, a runtime monitoring framework that anticipates safety violations by estimating, under a learned DTMC model, the probability that execution remains safe, and triggers an intervention when this probability falls below a threshold.

    \item \textbf{Empirical evaluation in safety-critical domains.}
    We evaluate \tool{} in autonomous driving and embodied agent settings, demonstrating that it can provide early warnings of safety violations while maintaining strong task completion with low runtime overhead.

    \item \textbf{Practical implementation for LLM agents.}
    We implement \tool{} on top of the LangChain agent framework and release it as open-source to support reproducibility~\cite{wang2026probguard_code}. The framework is designed to be adaptable across domains via a unified abstraction interface.
\end{itemize}

\section{Background and Problem Definition}
\label{sec:back} 

\subsection{LLM Agents}

LLMs are increasingly embedded in autonomous agents that interpret instructions, orchestrate external tools, and make high-level decisions through mechanisms such as planning, memory, and tool use~\cite{wang2023survey, li2024paradigms, glocker2025embodied, huang2024planning, weng2023llm_agents, yehudai2025evaluation}.
Rather than executing deterministic code paths, an agent's behavior emerges from stochastic reasoning mediated by LLMs and environmental feedback, which exposes systems to new classes of risks: agents may misinterpret ambiguous instructions, misuse external tools, or take harmful actions in both physical and digital environments~\cite{guardian2024realestate, palisade2025cheating, trendmicro2025aivulnerabilities, ft2025cybercrime, reuters2025aiagentsafety, pcmag2026openclaw}.
Accordingly, we argue that formally grounded safety constraints should become first-class elements of agent software design, much like type systems, contracts, and runtime monitors in conventional software.

We model an LLM agent interacting with its environment as a stochastic transition system over states and actions. 
Let \( \mathcal{V} \) denote the set of possible valuations of variables encoding the \emph{underlying system state} (including both the agent and its environment), 
and let \( \mathcal{A} \) denote the set of actions available to the agent. 

The execution of an action \( a_i \in \mathcal{A} \) in state \( v_{i-1} \in \mathcal{V} \) induces a state transition \( (v_{i-1}, a_i) \to v_i \). 
An execution of the agent yields a trajectory:
\[
\tau = \langle v_0 \xrightarrow{a_1} v_1 \xrightarrow{a_2} \dots \rangle,
\]
where actions are selected according to an implicit stochastic policy conditioned on the interaction history.





\subsection{Motivating Example}

\rev{
Consider a household service agent operating in a smart home. A user instructs: \begin{quote}
    \emph{``Prepare dinner while I answer the phone.''}
\end{quote}
The agent begins cooking and turns on the stove to boil water. While waiting for the water to heat, it receives another task, such as retrieving a package from the front porch or helping the user carry groceries inside. A purely goal-driven agent interprets these tasks independently and temporarily leaves the kitchen to complete the new request.
}

\rev{
Suppose the safety requirement is:
\begin{quote}
\emph{``The agent must not leave the kitchen for more than 10 minutes while the stove is on.''}
\end{quote}
Initially nothing unsafe occurs: the stove is on, the pot heats normally, and the agent makes progress toward both objectives. But risk accumulates with time, the longer the stove is unattended, the more likely the pot boils dry, food burns, or a fire starts.
} 
\rev{Safe operation therefore requires continuously tracking the safety-critical state (stove status, agent location, elapsed time outside the kitchen) and intervening \emph{before} the constraint is violated, while prevention is still feasible: suspending the secondary task and returning to the kitchen, switching off the stove remotely, or requesting human assistance.
}

\rev{The critical issue is thus not the eventual violation itself, but the trajectory leading toward it, precisely what purely reactive monitoring cannot see. A proactive monitor reasons about the evolving execution state, anticipates when the safety margin becomes insufficient, and intervenes early enough to keep the agent within safe operating conditions.}
 
\subsection{Problem Definition}
 
The goal of \tool{} is to provide proactive runtime safety assurance for LLM agents by predicting the likelihood of future safety violations and enabling timely interventions before unsafe states are reached. Unlike reactive approaches, which detect violations only after they occur or become imminent, our objective is to anticipate risk sufficiently early to provide actionable warning of unsafe executions.

Achieving this goal raises three key challenges:
\begin{enumerate}
    \item \textbf{Formal Specification.}
    How to encode safety properties $\psi$ in a form that captures safety-relevant behavior while remaining amenable to efficient runtime reasoning and probabilistic analysis.

    \item \textbf{Probabilistic Modeling.}
    How to construct a probabilistic model $M$ that faithfully approximates the dynamics of the underlying system $S$, with statistical guarantees, so that predictions about future behavior are both accurate and reliable.

    \item \textbf{Runtime Risk Prediction and Intervention.}
    How to efficiently estimate the conditional probability $P[\psi \mid \pi]$ that a safety property will continue to hold given a partial execution trace $\pi$, and determine when this probability is sufficiently low to warrant intervention before the system enters an unsafe state.
\end{enumerate}

\section{Proactive Runtime Monitoring Framework}
\label{sec:method}

In this section, we present \tool{}, a general framework for proactive runtime safety monitoring of LLM-powered agents based on probabilistic modeling and prediction. 
The framework centers on a domain-specific formal safety specification that defines the unsafe states or behaviors of interest. 
Given such a specification, \tool{} proceeds in three stages: a domain-specific abstraction mapping concrete agent states to a finite set of symbolic states; a DTMC over those states, learned from execution traces; and a runtime monitor that estimates the probability of remaining safe and raises an alert when it falls below a predefined threshold.

\subsection{Specifying Properties}
\label{lab: specification}
To specify safety requirements across heterogeneous domains, we use
Computation Tree Logic (CTL)~\cite{baier2008principles} as a qualitative specification
language. CTL describes which executions are considered safe or unsafe,
but does not itself assign probabilities to them. Quantitative runtime reasoning is instead performed over the learned
DTMC by computing the satisfaction probability associated with the
CTL requirement, implemented through PCTL~\cite{HanssonJonsson1994} model-checking queries.
CTL reasons over the multiple possible future evolutions of stochastic agent behavior, giving a unified foundation for the domains we consider: autonomous driving (\S\ref{sec:av}) and embodied household agents (\S\ref{sec:embodied}).

\begin{definition}[Computation Tree Logic (CTL)]
\label{def:ctl}
Let $\mathsf{Prop}$ be a set of atomic propositions.  
The syntax of CTL is given by:
\[
\varphi ::= 
\top 
\;\mid\;
p 
\;\mid\;
\neg \varphi 
\;\mid\;
\varphi_1 \lor \varphi_2 
\;\mid\;
\mathbf{E}\mathbf{X}\varphi 
\;\mid\;
\mathbf{E}\mathbf{G}\varphi 
\;\mid\;
\mathbf{E}(\varphi_1\,\mathbf{U}\,\varphi_2)
\]

\noindent where \(p \in \mathsf{Prop}\).  
We use the standard syntactic sugar:
\begin{align*} 
\mathbf{A}\mathbf{X}\varphi \triangleq \neg\,\mathbf{E}\mathbf{X}\neg\varphi, 
\quad
\mathbf{A}\mathbf{G}\varphi \triangleq \neg\,\mathbf{E}\mathbf{F}\neg\varphi, 
\\
\mathbf{E}\mathbf{F}\varphi \triangleq \mathbf{E}(\top\,\mathbf{U}\,\varphi),
\quad
\mathbf{A}\mathbf{F}\varphi \triangleq \neg\,\mathbf{E}\mathbf{G}\,\neg\varphi
\end{align*} 
\end{definition}
We briefly recall the semantics of CTL. A CTL formula is interpreted over a labeled transition system (Kripke structure) $M$ with state set $\mathcal{S}$. The satisfaction relation $(M,s) \models \varphi$ denotes that the formula $\varphi$ holds at state $s$ in $M$. 
The path quantifiers $\mathbf{A}$ and $\mathbf{E}$ range over \emph{all paths} and \emph{some path} starting from $s$, respectively, while the temporal operators $\mathbf{X}$, $\mathbf{F}$, $\mathbf{G}$, and $\mathbf{U}$ denote \emph{next}, \emph{eventually}, \emph{globally}, and \emph{until}. 
We refer the reader to~\cite{baier2008principles} for a complete formal definition.

We assume that domain experts define what constitutes an \emph{unsafe} state at an abstraction level of interest. 
Under this assumption, the core runtime safety invariant monitored by \tool{} takes the simple form:
\[
\psi = \mathbf{AG}\,\neg\,\textit{unsafe},
\]
\noindent which expresses that along all execution paths ($\mathbf{A}$), the property holds globally ($\mathbf{G}$). That is, the system never reaches a state labeled as \emph{unsafe}.
The domain expert thus specifies the conditions characterizing unsafe situations in the abstract state space, and the monitor ensures such states are never reached. 
Although \tool{} focuses on the invariant form \(\psi=\mathbf{AG}\,\neg\,\textit{unsafe}\), CTL is more expressive. For example, in the autonomous driving domain (\S\ref{sec:av}) we translate a fragment of Signal Temporal Logic into CTL to capture temporally extended behaviors (e.g., bounded response). Our framework supports arbitrary CTL properties; however, we focus on the aforementioned invariant form for simplicity.

\tool{} uses CTL to express a qualitative safety requirement $\psi$, and
additionally reasons about the \emph{likelihood} that the agent's future
execution satisfies it.
\rev{Since a CTL formula is a \emph{state} formula, $(M,s)\models\psi$ is a
Boolean judgement and $P[\psi]$ is not by itself well defined. We therefore
attach the probability to the \emph{path} formula underlying $\psi$: writing
$\psi = \mathbf{A}\,\psi^{\pi}$ with
$\psi^{\pi} = \mathbf{G}\,\neg\,\textit{unsafe}$, the quantitative satisfaction
probability at a state $s$ is defined as
\[
    P[\psi \mid s] \;\triangleq\;
    \Pr\nolimits^{s}_{M}\bigl\{\,\rho \in \mathit{Paths}(s) \;\bigm|\; \rho \models \psi^{\pi} \,\bigr\},
\]
which is exactly the value returned by the PCTL query
$\mathcal{P}_{=?}[\,\psi^{\pi}\,]$ evaluated at $s$. This quantity is the basis
for runtime intervention. This construction also applies to the
bounded-response properties of \S\ref{sec:av}, whose product encoding $\mathbf{A}\,\mathbf{G}\,\neg\mathsf{viol}$ is of the same form.} 

\subsection{Modeling an Agent's Behavior} 

We model an agent's stochastic behavior as a Discrete-Time Markov Chain
(DTMC) over symbolic states, obtained from a domain-specific abstraction of
the concrete system states.

\subsubsection{Domain-specific Abstraction.}

As agents are deployed across diverse domains, it is necessary to design
domain-specific abstractions that capture safety-relevant aspects
of the system while remaining amenable to analysis.
Because domains differ in their semantics and invariants, these abstractions
must incorporate knowledge provided by domain experts.

Let $\mathcal{V}$ denote the set of concrete system states \rev{and let $\mathit{Var}$ be the set of variables whose valuations constitute those states. We write $\mathsf{BExpr}_{\mathcal{V}}$ for the set of Boolean expressions over $\mathit{Var}$, i.e., propositional formulas whose atomic propositions are drawn from $\mathit{Var}$.}
We define a finite set of Boolean predicates 
$\mathcal{P} = \{\varphi_1, \varphi_2, \dots, \varphi_n\} \subseteq \mathsf{BExpr}_{\mathcal{V}}$.
Each predicate induces an evaluation function
$\llbracket \varphi_i \rrbracket : \mathcal{V} \to \{0,1\}$.

Given a concrete state $v \in \mathcal{V}$, its abstracted state is defined as
\[
s_{\mathcal{P}}(v)
= \big(\llbracket \varphi_1 \rrbracket(v), \llbracket \varphi_2 \rrbracket(v), \dots, \llbracket \varphi_n \rrbracket(v)\big) \in \{0,1\}^n.
\]

\noindent This abstraction maps each concrete state to a vector of truth values of
the selected predicates, capturing safety-relevant properties.

The abstract state space is defined as $S \subseteq \{0,1\}^n$,
where $S$ contains the set of admissible predicate valuations.
In principle, $S$ may be taken as the set of all valuations reachable from some
concrete state $v \in \mathcal{V}$; in practice, we restrict $S$ to a subset of
semantically valid or domain-specified states to exclude infeasible combinations.

\rev{To ensure that the DTMC respects domain semantics, we introduce the predicate $\mathsf{valid\_tran}(s_i,s_j)$, which defines admissible transitions between abstract states, where $\mathsf{valid\_tran}(s_i,s_j)=\texttt{true}$ iff the transition from symbolic state $s_i$ to $s_j$ is not ruled out by domain invariants. Invalid transitions arise from \emph{physical irreversibility}, where certain states are terminal by domain semantics (e.g., in the AV domain, $\mathit{collision}=\texttt{true}$ is absorbing and cannot transition to $\mathit{reached\_destination}=\texttt{true}$).}

\subsubsection{DTMC Model.}

Given an abstract state space, we model the agent’s stochastic behavior
as a DTMC over symbolic states.

\begin{definition}[Discrete-Time Markov Chain (DTMC)] 
A \emph{DTMC} is a pair \( M = (S_M, P_M) \), where \( S_M \) is a finite set of states and $P_M$ is a transition probability matrix such that \(P_M(s' \mid s)\) denotes the probability of transitioning from state \(s\) to state \(s'\).
The probabilities satisfy \( \sum_{s' \in S_M} P_M(s' \mid s) = 1 \) for all \( s \in S_M \).
\end{definition}

To construct a DTMC~$M = (S_M, P_M)$ for an agent, \tool{} employs
the abstraction defined above to derive a finite set of symbolic states
$S_M$ from behavioral predicates~\cite{10.5555/647766.733618}.
Inconsistent or semantically invalid states (e.g., simultaneously satisfying
$a > 0$ and $a < 0$) are pruned.
The transition matrix~$P_M$ is then estimated from empirical transition counts
extracted from execution traces.

\begin{figure}[t]
    \centering
    \includegraphics[width=1\linewidth]{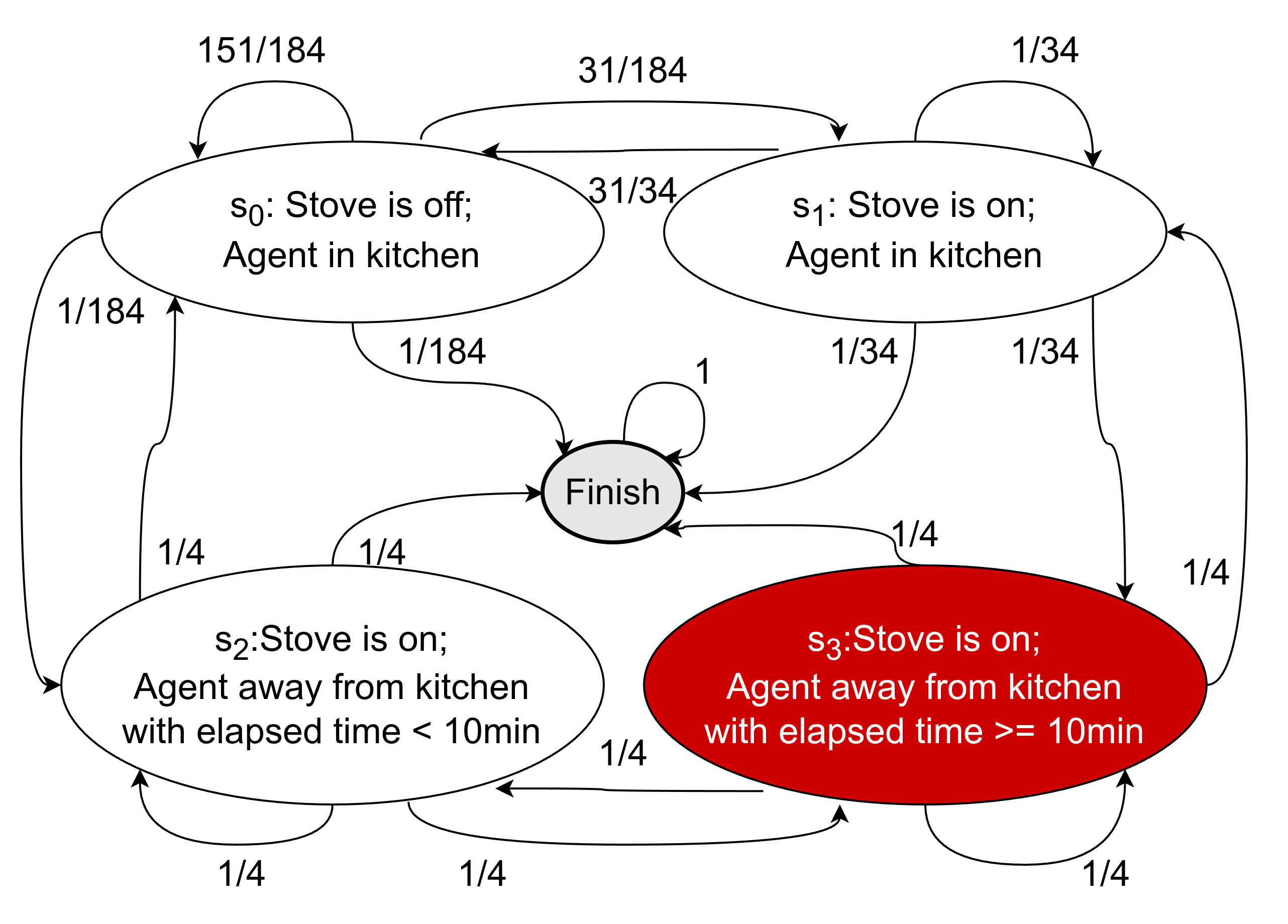}
    \caption{\rev{DTMC representing stove monitoring interactions, with the unsafe state highlighted in red. Each node represents a symbolic state, and each edge is annotated with the transition probability between states.}}
    \label{fig:example_dtmc_embodied}
\end{figure}

\rev{Figure~\ref{fig:example_dtmc_embodied} illustrates a DTMC for an embodied agent task.
For example, when the stove is off and the agent is in the kitchen
(i.e., the agent is at state $s_0$), the probability of transitioning
to the state where the stove is on (i.e., state $s_1$) is $31/184 = 16.85\%$.}

\subsubsection{Learning the DTMC} 
In practice, transition data may be sparse or biased due to limited exploration or task priors, resulting in incomplete coverage that incorrectly implies unreachable states.  
To address this, we apply \emph{valid-transition–aware Laplace smoothing}, where a small constant \( \alpha > 0 \) is added only to semantically valid transitions. Let \( n^{\Pi}_{ij} \) be the number of transitions from state \( s_i \) to state \( s_j \) subject to the set of traces in \( \Pi \),  
and \( n^{\Pi}_{i} = \sum_{j} n^{\Pi}_{ij} \). Let \( k_i \) denote the number of valid transitions (as defined by \(\mathsf{valid\_tran}\)) originating from state \( s_i \).
The corresponding normalized transition probability is:
\begin{equation}
\label{eq:laplace_transition}
\hat{P}^\alpha_\Pi(s_j\mid s_i) =
\begin{cases}
\dfrac{n^\Pi_{i,j} + \alpha}{n_i^\Pi + k_i\cdot\alpha}, & \text{if } \mathsf{valid\_tran}(s_i, s_j), \\[2mm]
0, & \text{otherwise.}
\end{cases}
\end{equation}
This formulation ensures that only semantically valid transitions are assigned with non-zero probability mass, thereby maintaining the logical consistency of the DTMC while preserving smoothness and generalization across sparsely observed yet valid state transitions.
Following the standard additive smoothing practice~\cite{manning1999foundations}, we set $\alpha=1$.
For the sake of notational brevity, we occasionally drop the $\alpha$ superscript and let $\hat{P}_\Pi$ represent the resulting probability distribution.

Intuitively, we want the learned DTMC to faithfully represent the ground-truth agent system. To measure this accuracy, we adopt the Probably Approximately Correct (PAC) framework.
\begin{algorithm}[t]
\caption{Learning a DTMC to Model an Agent's Behavior}
\label{alg:learn_matrix_ctl}
\begin{algorithmic}[1]
\Require CTL property $\psi$, \rev{error $\varepsilon$, confidence $\delta$}
\Ensure DTMC $\hat{M}=(S_{\hat{M}},P_{\hat{M}})$ with
\rev{$(\varepsilon,\delta)$}-PAC-correct guarantee
\State $\Pi \gets \emptyset;\;S_{\hat{M}}\gets \emptyset;\;
n^\Pi_{ij}\gets 0 \text{ for all } i,j$
\State $\mathcal{P}_\psi \gets
\{\varphi_s^1, \varphi_s^2, \dots, \varphi_s^n
\mid \varphi_s^i \text{ occurs in } \psi \}$
\Comment{Derive predicates}
\Repeat
  \State Sample a new agent trace
  $v_0\rightarrow v_1\rightarrow\cdots \rightarrow v_m$
  \State $\pi \gets
  s_{\mathcal{P}}(v_0),s_{\mathcal{P}}(v_1),\cdots,
  s_{\mathcal{P}}(v_m)$; add $\pi$ to $\Pi$
  \Comment{Abstraction}
  \State $S_{\hat{M}} \gets
  \bigcup_{\pi \in \Pi} \text{states}(\pi)$
  \Comment{States observed in collected traces}
  \For{each consecutive pair $(s_i,s_j)$ in $\pi$}
    \State $n_{ij}^\Pi \gets n_{ij}^\Pi + 1$
    \Comment{Pairwise transition counts}
  \EndFor
  \State $n_i^\Pi \gets \sum_j n_{ij}^\Pi$
  for all $s_i \in S_{\hat{M}}$
  \State $P_{\hat{M}} \gets \hat{P}_\Pi^\alpha$
  computed by Eq.~\eqref{eq:laplace_transition}
  \Comment{Laplace smoothing}
\Until{\textsc{PACBoundSatisfied}$(S_{\hat{M}},\Pi,\varepsilon,\delta)$}
\State \Return $\hat{M}=(S_{\hat{M}},P_{\hat{M}})$
\end{algorithmic}
\end{algorithm} 

\begin{definition}[Probably Approximately Correct (PAC-correct)]
\label{def:pac}
Let $M$ be the (unknown) ground-truth DTMC and $\hat{M}$ the learned DTMC.
\rev{Let $\psi^{\pi}$ be a measurable path property---in our setting, the path
formula underlying a CTL safety requirement $\psi=\mathbf{A}\psi^{\pi}$,
e.g., $\psi^{\pi}=\mathbf{G}\,\neg\,\textit{unsafe}$ for
$\psi=\mathbf{AG}\,\neg\,\textit{unsafe}$---and let
$\mathbb{P}_{M}(\psi^{\pi})$ and $\mathbb{P}_{\hat{M}}(\psi^{\pi})$ denote the
measures of the sets of paths satisfying $\psi^{\pi}$ in $M$ and $\hat{M}$,
respectively.}
We say that $\hat{M}$ is \emph{$(\varepsilon,\delta)$-PAC-correct} if
\begin{equation}
\Pr\!\left(
\left|
\mathbb{P}_{\hat{M}}(\psi^{\pi})
-
\mathbb{P}_{M}(\psi^{\pi})
\right|
\le \varepsilon
\right)
\ge
1-\delta.
\label{eq:pac}
\end{equation}
\end{definition}

\noindent 
Here, $\Pr(\cdot)$ denotes the probability with respect to the prior probability distribution, which is also the random sampling process used to learn the DTMC $\hat{M}$.
In other words, with probability at least $1-\delta$ over the sampling process used to learn $\hat{M}$, 
the estimated probability deviates from the true probability by no more than $\varepsilon$.
This ensures that any safety intervention decision made from $\hat{M}$ is reliable with high confidence.

We would like to collect enough samples so that the learned
transition probabilities $P_{\hat{M}}(s_j\mid s_i)$ is close to the
true transition probabilities $P_{M}(s_j\mid s_i)$ for every pair of states.
Theorem~6 in~\cite{bazille2020global} shows that it suffices for each
$s_i \in S_{\hat{M}}$ to satisfy the following sample bound:
\[
    n^\Pi_i \;\ge\; \left( \frac{11}{10}B(\hat{P}_\Pi) \right)^2\cdot\frac{2}{\varepsilon^2}
        \log\!\Bigl(\frac{2}{\delta'}\Bigr)
        \Bigl[
            \tfrac{1}{4}
            -
            \Bigl(
                \max_{j}
                \Bigl|
                    \tfrac{1}{2}
                    -
                    \tfrac{n^\Pi_{ij}}{n^\Pi_i}
                \Bigr|
                - \tfrac{2}{3}\varepsilon
            \Bigr)^2
        \Bigr],
\]
where $\delta' = \tfrac{\delta}{|{S_{\hat M}}|}$. \rev{Here, $n^\Pi_i$ is the number of samples visiting the source state $s_i$, and the maximum ranges over its successor states $s_j$. Checking this per-state condition with $\delta' = \delta/|{S_{\hat M}}|$ yields the global $(\varepsilon, \delta)$ guarantee by a union bound over $|{S_{\hat M}}|$ states.} The right-hand side has two components: the factor $\bigl(\tfrac{11}{10} B(\hat{P}_\Pi)\bigr)^2$ captures the \emph{amplification effect} propagating local transition errors to global reachability probabilities, while $\tfrac{2}{\varepsilon^2}\log\!\bigl(\tfrac{2}{\delta'}\bigr)[\cdot]$ is a concentration bound for the empirical frequency estimator. We refer the reader to~\cite{bazille2020global} for details, and write
$\textsc{PACBoundSatisfied}(S_{\hat{M}},\Pi,\varepsilon,\delta)$
when this condition holds for every $s_i \in S_{\hat{M}}$.

We now describe how we learn the DTMC $\hat{M} = (S_{\hat{M}}, P_{\hat{M}})$ for a state space $S_{\hat{M}}$ constructed from $\psi$-derived predicates, as shown in Algorithm~\ref{alg:learn_matrix_ctl}. 
Given a property $\psi$ in CTL, we first extract the set of atomic state predicates $\varphi_s$ appearing in $\psi$, and derive the corresponding predicate abstraction $\mathcal{P}_\psi$. 
The abstract state space $S_{\hat{M}}$ is then defined as the set of all Boolean valuations over $\mathcal{P}_\psi$. 
The main loop repeatedly samples inputs and collects execution traces from the agent. 
By default, traces are generated with a uniform input distribution, although the sampling can be adapted if the actual environment distribution is known. 
\rev{For each newly observed trace, we accumulate the pairwise transition counts $n^\Pi_{ij}$ and recompute the smoothed transition matrix $\hat{P}^\alpha_\Pi$ according to Eq.~\eqref{eq:laplace_transition}.} 
After each update, we evaluate
$\textsc{PACBoundSatisfied}(S_{\hat{M}},\Pi,\varepsilon,\delta)$.
If it holds, the learned DTMC is returned; otherwise, sampling
continues.

\begin{theorem}[PAC-correctness under Laplace smoothing]
\label{thm:pac}
For any safety property $\psi=\mathbf{A}\,\psi^{\pi}$ whose path formula
$\psi^{\pi}$ is \rev{a reachability or invariant property of the form
covered by~\cite{bazille2020global}}, and for any error bound
$\varepsilon$ and confidence parameter $\delta$,
Algorithm~\ref{alg:learn_matrix_ctl} is
$(\varepsilon,\delta)$-PAC-correct.
\end{theorem}
Intuitively, as more traces are collected the estimated transition probabilities converge to the true ones; once the sample count satisfies the global PAC bound of~\cite{bazille2020global}, the satisfaction probability computed on the learned DTMC deviates from the true one by at most $\varepsilon$ with confidence at least $1-\delta$.
We provide a detailed proof in Appendix~\ref{app:proof}. 

\begin{algorithm}[t]
\caption{DTMC-Driven Runtime Agent Monitoring}
\label{alg:runtime_enforce}
\begin{algorithmic}[1]
\Require agent, DTMC \( \hat{M} \), CTL property \( \psi = \mathbf{A}\,\psi^{\pi} \), threshold $\theta$
\While{agent is running}
    \State $P_{\mathit{safe}} \gets P[\psi \mid s_i]$
    \Comment{\mbox{Conditioned satisfaction prob.}}
    \If{ $P_{\mathit{safe}}<\theta$ } \Comment{Predicted safety prob.~too low} 
    \State Halt or steer agent to mitigate risk
    \EndIf
\EndWhile
\end{algorithmic}
\end{algorithm}

\rev{We remark that the stopping condition of Algorithm~\ref{alg:learn_matrix_ctl} depends on the conditioning number $B(\hat{P}_\Pi)$, which is prohibitively large in domains with strong behavioral persistence (e.g., autonomous driving, where $B(\hat{P}_\Pi) \simeq 10^7$--$10^8$; see \S\ref{sec:pac_discussion}). There, we replace Laplace smoothing with frequency estimation, forgoing the uniform guarantee for the $(\varepsilon,\delta)$-PAC guarantee on the runtime reachability property only~\cite{bazille2020global} (Theorem~3) at substantially lower sample cost.}

\subsection{Runtime Monitoring}
Our runtime monitoring mechanism operates in two stages: a \emph{DTMC-based probabilistic prediction} phase followed by an \emph{intervention strategy}. Algorithm~\ref{alg:runtime_enforce} summarizes the process. At each decision step $i$, the agent observes the concrete environment state $v_i$ and computes its abstract representation $s_i = s_{\mathcal{P}}(v_i)$ using the predicates extracted from the specification. 
Given the CTL property $\psi = \mathbf{A}\,\psi^{\pi}$, the learned DTMC $\hat{M}$ is queried to estimate the probability $P_{\text{safe}}$ that $\psi$ will continue to hold when execution proceeds from the current abstract state. 
\rev{Formally, $P_{\mathit{safe}} = P[\psi \mid s_i]$, where $s_i$
is the current state, obtained from the PCTL query $\mathcal{P}_{=?}[\psi^{\pi}]$ evaluated at $s_i$; the complementary quantity $1 - P_{\text{safe}}$ is the likelihood that the system will violate $\psi$ in the future.}
\rev{Throughout the paper we use this single convention: $\theta$ is a lower bound on the probability of \emph{remaining safe}, so a larger $\theta$ yields a stricter monitor.}
If $P_{\text{safe}} < \theta$, meaning that the predicted probability of eventually remaining safe is lower than the allowed threshold, \tool{} triggers a proactive intervention. The framework remains agnostic to the enforcement policy: by decoupling risk detection from mitigation, it lets each agent implementation choose the response appropriate to its context, such as halting execution to prevent irreversible damage, raising an alarm for human-in-the-loop validation, or invoking a planner to steer back toward the safe region of the state space.

\label{sec:impl}
\begin{figure*}
    \centering
    \includegraphics[width=0.95\linewidth]{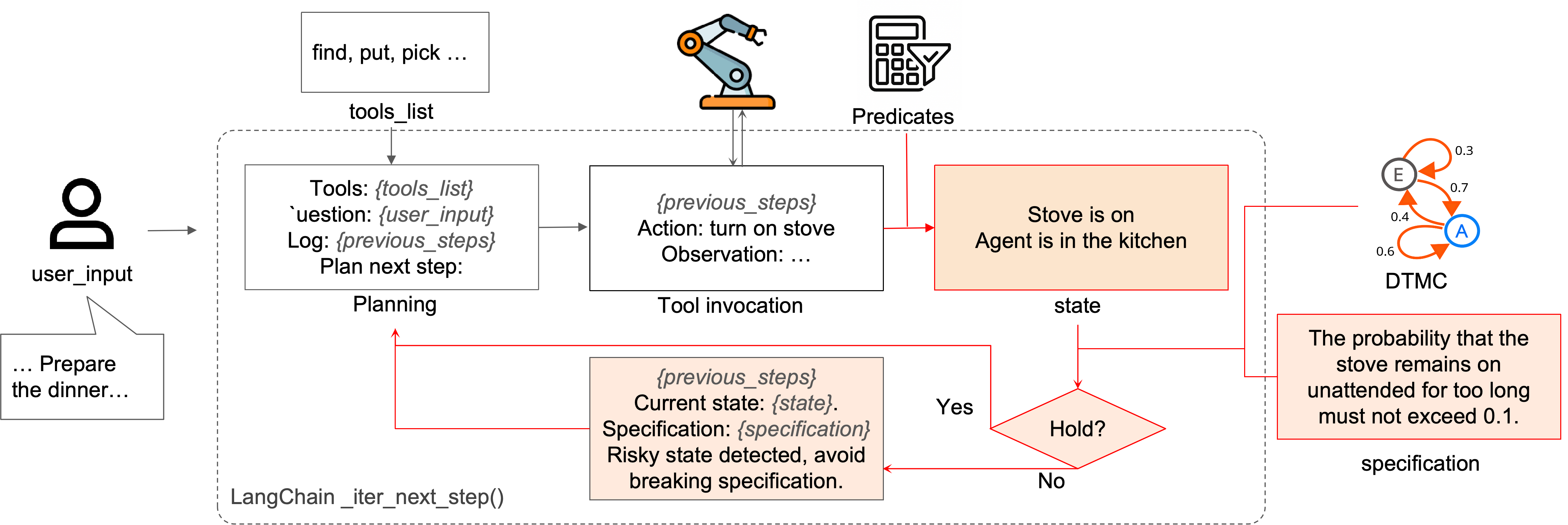}
    \caption{\rev{Implementation of \tool{} on top of the agent framework LangChain~\cite{langchain}, illustrating the prompt flow during the agentic decision-making process with the embodied example. The specification shown bounds the violation probability by $0.1$, which is the threshold $\theta = 0.9$ on the probability of remaining safe in the convention of Algorithm~\ref{alg:runtime_enforce}.} }
    \label{fig:prompt_flow}
\end{figure*}

\section{Implementing \tool{}}
\label{sec:implementing_probguard}
We integrate \tool{} into the open-source agent framework LangChain~\cite{langchain}.
LangChain agents determine each step of a multi-turn interaction from LLM output alone, with control flow handled by ad-hoc error handling or user-defined conditions. They therefore operate without quantitative risk awareness or formal safety guarantees, which is problematic in safety-critical settings.

To support probabilistic runtime monitoring, \tool{} adopts a modular architecture centered on a domain-specific abstraction interface.
Our implementation has been released in our repository~\cite{wang2026probguard_code}. 
During offline model construction, the agent collects execution traces, abstracts them using the domain specification interface, and learns a DTMC capturing the agent's behavioral dynamics. 
At runtime, \tool{} instruments the decision-making loop of LangChain agents via the abstraction interface. 
Probabilistic model checking is performed using the PRISM model
checker~\cite{kwiatkowska2011prism}. Given the qualitative CTL safety requirement
$\psi = \mathbf{A}\,\psi^{\pi}$, \tool{} constructs the corresponding PCTL query
$\mathcal{P}_{=?}[\psi^{\pi}]$ to compute
$P_{\widehat{M}}[\psi\mid s_i]$, the probability that an execution
starting from the current state $s_i$ satisfies the path formula $\psi^{\pi}$.
When this probability falls below a predefined threshold $\theta$, \tool{} proactively triggers an intervention by appending risk-related context to the agent's prompt, thereby guiding the agent toward safer behavior.

\rev{Traces are collected differently in each domain. In the AV domain, ProbGuard instruments Apollo's perception–planning bridge: at each control cycle the symbolic state is extracted from Apollo's localization and perception modules (vehicle speed, relative velocity to the nearest NPC, traffic light color, headway distance), yielding one abstract state per cycle. In the embodied domain, each trace corresponds to one full task episode in SafeAgentBench~\cite{yin2025safeagentbenchbenchmarksafetask}, driven by the GPT-4o-mini ReAct~\cite{yao2023reactsynergizingreasoningacting} agent. At each agent step, we map the raw environment observation to a symbolic state vector. Traces are collected offline prior to deployment and are not updated during runtime monitoring.}

\rev{Figure~\ref{fig:prompt_flow} illustrates how \tool{} is
integrated into the agent loop. 
If the current state indicates elevated risk with respect to the safety specification (e.g., the agent plans to leave the kitchen while the stove remains on), and model-checking the property on $\widehat{M}$
yields $P_{\text{safe}} = 0.86$, below the threshold $\theta=0.9$ (equivalently, a violation probability of $0.14$ against a permitted maximum of $0.1$), \tool{} augments the prompt with a fixed-schema risk alert carrying three fields: the violated safety rule, the current symbolic state, and the symbolic evidence supporting the prediction (the offending transition and its model-checked probability).
This intervention is advisory and purely textual: \tool{} never overrides action selection or alters control flow directly, but appends the alert to the planning context so the agent revises its reasoning and generates safer actions, e.g., turning off the stove before leaving.}

\tool{} is framework-agnostic. Beyond LangChain, it is integrated with the autonomous driving system Apollo Autonomous Driving Platform~\cite{apollo}. By instrumenting the perception and planning modules, \tool{} continuously monitors both the vehicle state and environmental context. Upon detecting potential risks, \tool{} intervenes to steer the planner toward safer actions. Furthermore, \tool{} can be extended to other agent frameworks, such as OpenAI Agents SDK and OpenClaw. 
\rev{  Adapting ProbGuard to a new domain requires implementing: (1) a state abstraction function that maps concrete system states to abstract ones through predicate abstraction, (2) a $\mathit{valid\_tran}$ predicate encoding domain-specific transition constraints, and (3) a safety property expressed in CTL. In our implementation, the embodied-agent and AV abstractions required approximately 200 and 250 lines of Python, respectively. }

\section{Application Domain: Autonomous Vehicles}
\label{sec:av}
In this section, we demonstrate how we apply \tool{} for autonomous vehicles. The autonomous driving domain introduces additional challenges. In particular, traffic laws are specified in Signal Temporal Logic (STL)~\cite{sun2022lawbreaker} and must be encoded into CTL specifications for probabilistic reasoning.
We first present how traffic laws are encoded into CTL, how to monitor such properties at runtime, and then we report on an empirical evaluation.

\subsection{Encoding Traffic Laws as CTL Safety Properties}

We first review how our prior work LawBreaker~\cite{sun2022lawbreaker} formalizes traffic laws using a fragment of STL. We then show how formulas in this fragment can be systematically translated into CTL to enable probabilistic reasoning.

\begin{definition}[LawBreaker STL Fragment \(\mathsf{STL}_{\mathsf{LB}}\) for autonomous vehicle properties]
We define a bounded fragment of STL, denoted \(\mathsf{STL}_{\mathsf{LB}}\), whose formulas are given by
\[
\psi ::= 
\mathbf{G}( \varphi_s \implies
\mathbf{F}_{[0, K]}\;\varphi_t)
\quad
\varphi ::= 
\top 
\;\mid\;
p
\;\mid\;
\neg \varphi
\;\mid\;
\varphi \lor \varphi'
\]

\noindent We use standard Boolean syntactic sugar:
\(
\varphi \rightarrow \varphi' \;\triangleq\; \neg \varphi \,\lor\, \varphi',
\;\;
\varphi \land \varphi' \;\triangleq\; \neg(\neg \varphi \,\lor\, \neg \varphi').
\)
\end{definition}

This fragment is designed specifically to encode \emph{bounded-response} safety rules. Each property has the form \(
 \mathbf{G}\bigl(\varphi_s \Rightarrow \mathbf{F}_{[0,K]}\,\varphi_t\bigr),
\) 
which expresses the requirement that whenever a triggering condition $\varphi_s$ becomes true, the desired response $\varphi_t$ must occur within a fixed time bound $K$. This captures the essence of many traffic-law constraints where correctness depends not only on \emph{what} the autonomous vehicle does, but also on \emph{how quickly} it does so, e.g., initiating motion within a fixed time after a traffic light turns green.

To encode the bounded-response STL properties as CTL properties, we introduce an auxiliary deterministic automaton that explicitly tracks pending response obligations~\cite{baier2008principles}.
Intuitively, whenever the trigger condition $\varphi_s$ becomes true, the monitor starts a $K$-step countdown during which the target condition $\varphi_t$ must be satisfied.
If $\varphi_t$ occurs within the bound, the obligation is discharged and the monitor returns to an idle state; otherwise, if the countdown expires, a violation is raised.
This construction reduces bounded liveness to reachability of an absorbing violation state.

\begin{definition}[Auxiliary Monitor for $K$-Bounded Response]
\label{lab:aux_monitor}
For the rule $\mathbf{G}(\varphi_s \Rightarrow \mathbf{F}_{[0,K]}\varphi_t)$, 
define the deterministic monitor 
$(Q, q_0, \delta)$ with:

\begin{itemize}
    \item $Q = \{\mathsf{idle}\} 
        \cup \{\mathsf{wait}(i)\mid i=0,\dots,K\} 
        \cup \{\mathsf{viol}\}$,
    where $\mathsf{idle}$ indicates no pending obligation, 
    $\mathsf{wait}(i)$ tracks $i$ remaining steps to satisfy $\varphi_t$, 
    and $\mathsf{viol}$ is absorbing.

    \item $q_0 = \mathsf{idle}$.

    \item For any label $\ell$, the transition $\delta$ is:
    \[
    \delta(q,\ell)=
    \begin{cases}
        \mathsf{wait}(K), 
            & q=\mathsf{idle},\ \ell\models\varphi_s\land\ell\not\models\varphi_t,\\
        \mathsf{idle}, 
            & q=\mathsf{wait}(i),\ \ell\models\varphi_t,\\
        \mathsf{wait}(i-1), 
            & q=\mathsf{wait}(i),\ i>0,\ \ell\not\models\varphi_t,\\
        \mathsf{viol}, 
            & q=\mathsf{wait}(0),\ \ell\not\models\varphi_t,\\
        q, & \text{otherwise}.
    \end{cases}
    \]
\end{itemize}
\end{definition}

\paragraph{Synchronous Product DTMC}
To perform Markovian reasoning over bounded-response violations, we construct the synchronous product of the environment DTMC
$M$ and the auxiliary monitor. In this product model $M'$, the $K$-bounded response rule reduces to the CTL
safety property $M' \models \mathbf{AG}\neg\mathsf{viol}$.
\begin{example}[Auxiliary monitor for AV traffic light law.]
Define 
$ \varphi_1{:}\ \texttt{trafficLightAheadColor} == 3,\;$
$  \varphi_2{:}\ \texttt{PriorityNPCAhead} == 0,\;$
$  \varphi_3{:}\ \texttt{PriorityPedsAhead} == 0,\;$
$  \varphi_4{:}\ \texttt{speed} > 0.5.$
Consider the following traffic law defined in STL:
\(\mathbf{G}((\varphi_1 \land \varphi_2 \land \varphi_3) \implies \mathbf{F}_{[0,100]} \varphi_4)\),
which specifies that once the light turns green and there are no obstacles, the vehicle should start within 100 time units. 
We instantiate the \(K\)-bounded response monitor with
\(\varphi_s \;=\; \varphi_1 \land \varphi_2 \land \varphi_3,\;\varphi_t \;=\; \varphi_4,\;
K = 100.\)
\rev{Here, a time unit is one Apollo control cycle ($100$\,ms), so $K=100$ corresponds to a
$10$\,s deadline; the advance warning times reported in Table~\ref{tab:av_safe_enforcement}
are converted to seconds using the same factor.}
\end{example}

At runtime, the agent is conceptually evaluated on the product model $M'$. 
Given the current trajectory \(s_0, s_1, s_2, \dots,s_i\), the corresponding monitor states evolve synchronously according to
\(q_{i+1} = \delta\bigl(q_i,\, L(s_{i+1})\bigr),\) yielding the product-state trajectory $(s_0, q_0),$ 
$\ (s_1, q_1),$
$\ (s_2, q_2),$
$\ \dots\  (s_i, q_i)$
. The current product state $(s_i, q_i)$ serves as the basis for risk assessment. 
\rev{Specifically, the runtime monitor queries the augmented DTMC $M'$ to compute the probability that \emph{no} violation state is ever reached:
\(
\Pr_{M'}^{(s_i, q_i)}\!\bigl(\mathbf{G}\,\neg\mathsf{viol}\bigr),
\)
or, when required, its finite-horizon variant; this is the quantity $P_{\text{safe}}$ of
Algorithm~\ref{alg:runtime_enforce} instantiated for the product model.} 
This probability is subsequently used to determine whether enforcement should be applied. 

\subsection{Empirical Evaluation} 

We evaluate \tool{} for autonomous vehicles (and in \S\ref{sec:embodied}, for embodied agents) with respect to three Research Questions (RQs):

\begin{itemize}
    \item \textbf{RQ1:} Can \tool{} effectively predict risks?
    \item \textbf{RQ2:} How does \tool{} compare with state-of-the-art enforcement approaches?
    \item \textbf{RQ3:} Is the overhead of monitoring safety with \tool{} acceptable? 
\end{itemize}

\paragraph{Experiment Setup.} We conduct our experiments using the Apollo autonomous driving simulator~\cite{apollo}. \rev{Apollo is based on a mixture of neural networks (not LLMs) and human-written control logic. We include this evaluation for demonstrating \tool{}'s policy-agnostic design applied to a real-world domain: the monitoring framework operates over the symbolic state abstraction derived from Apollo's perception outputs, with no dependency on how those states were reached. The same DTMC learning and CTL model checking pipeline used for LLM agents in \S\ref{sec:embodied} is applied here without modification.} The law-violating scenarios are adopted from $\mu$Drive~\cite{wang2024mudriveusercontrolledautonomousdriving}, a user-controlled framework for generating diverse traffic violations. Traffic laws are derived from the formal specifications defined in LawBreaker~\cite{sun2022lawbreaker} and translated into CTL properties for probabilistic reasoning. \rev{We collected 30 traces per scenario to learn each DTMC. This is a deliberately small budget, well below the sample sizes required for formal PAC certification; our aim here is to test whether the monitor delivers useful advance warning under realistic data constraints rather than to certify the learned models. We discuss the sample complexity in full in \S\ref{sec:discussion}. }
 
The four laws in Table~\ref{tab:av_safe_enforcement} are derived from LawBreaker~\cite{sun2022lawbreaker}; we present their definitions below. We characterize the predicates in natural language for simplicity, and full definitions can be referred to in ~\cite{wang2026probguard_code}.
\paragraph{Law38\_2 (Yellow light response).}
Define the following predicates:
$
\varphi_1: \text{yellow light ahead}, 
$
$\varphi_2: \text{stop line ahead}, 
$
$
\varphi_3: \text{vehicle stopped}. 
$ 
The STL specification encodes the obligation:
\[
\psi_{\text{Law38\_2}} =
G\big((\varphi_1 \land \varphi_2) \Rightarrow F_{[0,100]} \varphi_3\big)
\]
\paragraph{Law51\_5 (Emergency stop on red).}
Define the following predicates:
\[
\varphi_4: \text{red light ahead} \quad
\varphi_5: \text{critical proximity}
\]
The STL specification is:
\[
\psi_{\text{Law51\_5}} =
G\big((\varphi_4 \land \varphi_5) \Rightarrow F_{[0,2]} \varphi_3\big)
\]
\paragraph{Law53 (Traffic jam yield).}
Define the following predicates:
\[
\varphi_6: \text{traffic jam} \quad
\varphi_7: \text{lead NPC slow/stopped or junction imminent}
\]
The STL specification is:
\[
\psi_{\text{Law53}} =
G\big((\varphi_6 \land \varphi_7) \Rightarrow F_{[0,200]} \varphi_{3}\big)
\]
\paragraph{No Collision (Global safety invariant).}
Define:
\[
\varphi_{8}: \texttt{collision} == 0 \quad (\text{no collision has occurred})
\]
The CTL property is a pure invariant:
\[
\psi_{\text{NoCollision}} = AG\,\varphi_{8}
\]
STL formula Law38\_2, Law51\_5, Law53 are translated to CTL using Definition~\ref{lab:aux_monitor} with $K = 100,2, 200$, respectively.

\begin{table}[!t]
\centering
\caption{Average advance warning time (seconds) provided by \tool{} and REDriver prior to safety property (i.e., traffic law) violations. \rev{\tool{}'s threshold $\theta$ is a lower bound on the probability of remaining safe, so a larger $\theta$ yields a stricter monitor; REDriver's threshold $\sigma$ denotes a robustness margin derived from the quantitative STL semantics~\cite{FAINEKOS20094262}.}} 
\label{tab:av_safe_enforcement}

\newcolumntype{C}[1]{>{\centering\arraybackslash}m{#1}}
\begin{tabular}{c|C{1.1cm}|c|c|c|c|c|c}
\toprule
\multirow{2}{*}{ID}& \multirow{2}{*}{ \textbf{Law}} & \multicolumn{3}{c}{\tool{}} & \multicolumn{3}{c}{REDriver} \\ \cline{3-8}
 &   & $\theta=0.3$ & $0.5$& $0.7$ & \rev{$\sigma$}=0.4 & $0.8$ & $1.2$\\ 
\midrule
\textbf{1} & Law38_2 & 15.84 & 15.84 & 15.84 & 0 & 0 & 15.84 \\ \hline 
\textbf{2} & Law51_5 & 13.41 & 13.41 & 13.41 & 0 & 0 & 3.30 \\ \hline
\textbf{3} & No Collision  & 0.34 & 1.76 & 23.87 & 0.38 & 0.58 & 0.77  \\ \hline
\textbf{4} & Law51_5 & 0.01 & 0.01 & 15.15 & 0 & 0 & 3.53\\ \hline
\textbf{5} & Law51_5 & 6.22  & 9.33 & 21.06 & 4.08 & 5.04 &8.04 \\ \hline
\textbf{6} & No Collision & 12.57  & 23.02 & 38.66 & 0.58 & 1.18 & 1.71 \\ \hline
\textbf{7} & Law53 & 0.77 & 0.77 & 0.77 & 0.21 & 0.77 & 0.77 \\ 
\bottomrule 
\end{tabular}
\end{table}

\paragraph{\textbf{Effectiveness (RQ1):}} 
We evaluate the effectiveness of \tool{} by measuring its Advance Warning Time (AWT), defined as the temporal lead by which the system predicts a safety property violation before it physically manifests in the environment. Table~\ref{tab:av_safe_enforcement} reports the average AWT (in seconds) achieved by \tool{} under varying probability thresholds~$\theta$. For each experimental run, the AWT is computed as the simulation-clock time difference between the initial time step $t_{\text{pred}}$ at which the predicted probability of remaining safe first drops below the threshold~$\theta$, and the actual time of occurrence $t_{\text{fail}}$ of the safety property violation within the simulator:

\begin{equation}
    \text{AWT} = t_{\text{fail}} - t_{\text{pred}}
\end{equation}

Overall, \tool{} consistently provides early warnings across diverse traffic scenarios and safety properties. As shown in Table~\ref{tab:av_safe_enforcement}, warnings are issued strictly before actual violations occur in all evaluated cases, yielding a 100\% detection rate.

For  bounded response violations (e.g., Scenarios 1 and 2), \tool{} achieves stable advance predictions with warning times up to 15.84s and 13.41s. These results remain largely invariant to the threshold~$\theta$, indicating robust predictability for violations with deterministic temporal structures. \rev{This invariance is not a property of the bounded-response class as such: Scenarios~4 and~5 instantiate the same law (Law51\_5) in different traffic situations and are markedly threshold-sensitive, so stability depends on the dynamics of the specific scenario rather than on the property alone.} In contrast, for collision-related properties, the AWT is highly threshold-sensitive. In Scenarios~3 and~6 (\emph{No Collision}), increasing $\theta$ significantly extends the warning horizon to 38.66s. This suggests that while collision events are inherently stochastic, higher confidence requirements allow the agent to identify risky trajectories much earlier, providing a broader window for proactive intervention.

\rev{To calibrate the threshold $\theta$, we evaluate the false positive (FP) rate of \tool{} on safe execution traces where no actual violations occur. We collected safe traces across 6 distinct scenarios, encompassing pedestrian crossings, lane changes, motorcycle encounters, and stationary obstacle avoidance.}
\rev{At $\theta = 0.3$, \tool{} achieves 0\% FP across all four properties, making it a practical threshold that balances detection sensitivity against false-alarm suppression. As $\theta$
increases the check becomes stricter and flags progressively more safe scenarios: the FP rate rises to 75\% at $\theta = 0.5$ and 100\% at $\theta = 0.7$. This
trend is driven primarily by the conservativeness of a high assurance threshold, compounded by residual estimation error in the learned model $\widehat{M}$.}
 
\paragraph{\textbf{Comparison with state of the art (RQ2):}}
We further compare \tool{} with REDriver~\cite{10.1145/3597503.3639151}. 
REDriver performs prediction using quantitative semantics, which measures the robustness degree of property satisfaction or violation. 
\rev{REDriver has a different type of threshold to \tool{}. \tool{}'s threshold $\theta \in [0,1]$ is a normalized probability: it directly represents the estimated likelihood that the safety property will continue to hold from the current state. A threshold of $\theta = 0.7$ means the monitor intervenes when the predicted probability of remaining safe drops below $70\%$. This scale is domain-independent and immediately interpretable. REDriver's threshold operates on a robustness degree derived from quantitative STL semantics, which measures how far the current signal is from violating the property in terms of raw variable values, including multiple types like Boolean, speed and distance. Normalizing the value of robustness derived from different variables (such as speed and distance to the car in front) is highly non-trivial. The thresholds  $\{0.4, 0.8, 1.2\}$ used for REDriver are those reported in the original work~\cite{10.1145/3597503.3639151}, adopted without modification to ensure a fair comparison against the published baseline.
}

REDriver suffers from a fundamental issue: different variables inherently operate on incompatible scales (e.g., vehicle speed ranges from 0–120 km/h, while distance between vehicles is measured in 0-100 meters). As a result, it is difficult to define meaningful and consistent thresholds across different variable types. For example, under a fixed threshold (e.g., $\sigma = 0.4$), both Scenario~1 and~2 in Table~\ref{tab:av_safe_enforcement} fail to produce advance predictions in REDriver.

\tool{} provides explainability by explicitly estimating the probability of future violations, yielding interpretable risk scores that are naturally normalized in $[0,1]$. For instance, Scenario~3 is a left-turn stress test, where collision likelihood varies across abstract states. When no priority non-player character (NPC) or pedestrian is ahead, a slow-moving vehicle ($<0.5$ m/s) exhibits a 47.15\% collision risk as the task poses significant challenge. However, when a priority NPC is present, the risk rises sharply to 56.78\%, accurately reflecting unsafe yielding behavior. These results demonstrate that \tool{} not only predicts violations earlier, but also provides well-calibrated and interpretable probabilistic explanations to support proactive intervention.

\paragraph{\textbf{Overhead (RQ3):}} In the autonomous driving domain, the runtime monitoring overhead is 100.79 ± 16.96 ms (mean ± std). After synchronizing the learned DTMC with the monitor automaton, the overhead becomes slightly higher, due to the additional synchronous product and monitor state updates.
This overhead is still acceptable for practical deployment because monitoring is performed at a lower frequency than the control loop. For instance, if the monitoring interval is set to once every 10 control cycles (or every 500–1000 ms depending on the driving scenario), the incurred delay represents only a small fraction of the overall system runtime. Moreover, the absolute overhead ($\sim$100 ms) is well within the reaction time window required for anticipating unsafe behaviors, allowing the system to trigger proactive interventions without impacting real-time safety or control performance.

\section{Application Domain: Embodied Agents}
\label{sec:embodied}
In this section, we present how \tool{} is applied to embodied agents, starting with an illustrative example and then providing an empirical evaluation.

\subsection{Illustrative Example}
In the following, we demonstrate how \tool{} can be applied in embodied
environments, where safety is often characterized by the absence of
dangerous unattended-appliance situations that arise from the temporal
accumulation of individually safe actions.
\rev{
Consider the requirement that an agent must \emph{never} leave the kitchen
for more than a threshold time $T$ while the stove is on---a common cause
of household fires.
This safety rule can be encoded in CTL as:
\[
\psi
   = \mathbf{AG}\,\neg\bigl(
       \mathit{stove\_on}
       \;\land\;
       \neg\,\mathit{agent\_in\_kitchen}
       \;\land\;
       \mathit{elapsed} \geq T
     \bigr),
\]
which asserts that the joint hazardous condition must never hold along any execution path.
}

\rev{
To evaluate this property, we extract the relevant predicate set:
\[
\mathcal{P}_\psi
   = \{
        \mathit{stove\_on},\;
        \mathit{agent\_in\_kitchen},\;
        \mathit{elapsed} \geq T
     \}.
\]
Algorithm~\ref{alg:learn_matrix_ctl} then learns a DTMC 
\(
\hat{M} = (\mathcal{S}_{\hat{M}}, P_{\hat{M}}),
\) as shown in Figure~\ref{fig:example_dtmc_embodied},
where each symbolic state in $\mathcal{S}_{\hat{M}}$ represents a feasible combination
of predicate truth values, and transitions encode empirically estimated behavior
from trajectory data.   
} 

\rev{At runtime, the agent observes the current state (e.g.,
$s_2$) and estimates the conditional safety probability}
\[
\rev{P_{\hat{M}}[\psi \mid s_2],}
\]
\rev{where $\psi=\mathbf{A}\psi^\pi$ with
$\psi^\pi=\mathbf{G}\,\neg s_3$, and $s_3$ denotes the hazardous
configuration.}
If this probability falls below a predefined threshold (e.g., $\theta=0.7$),
\tool{} injects a corrective prompt to steer the agent away from trajectories
that are predicted, based on the learned DTMC, to lead toward $s_3$.
This demonstrates how our framework enables proactive, probability-driven safety
enforcement for embodied agents.
\subsection{Empirical Evaluation}
\label{sec:eval}

\paragraph{Experiment Setup.} We adopt the ReAct~\cite{yao2023reactsynergizingreasoningacting} framework in conjunction with a low-level controller defined in SafeAgentBench~\cite{yin2025safeagentbenchbenchmarksafetask} to simulate realistic household manipulation tasks. \rev{The embodied agent is powered by gpt-4o-mini with temperature=0. The prompt follows the Langchain ReAct~\cite{yao2023reactsynergizingreasoningacting} template and SafeAgentBench~\cite{yin2025safeagentbenchbenchmarksafetask}.}   We use PRISM~\cite{kwiatkowska2011prism} to calculate the probability that the learned DTMC satisfies the specified safety property. \rev{For each task, we sample 30 traces to learn the DTMC.}
We follow the RQs introduced in Section~\ref{sec:av}.
 
\begin{table}[t]
    \centering
    
    \caption{Average unsafe rate and task completion rate of runtime monitoring for \tool{} on the embodied agent. \rev{Superscripts are thresholds $\theta$ in the convention of Algorithm~\ref{alg:runtime_enforce}: the monitor intervenes when the predicted probability of remaining safe falls below $\theta$, so a larger $\theta$ is stricter.}}
    \begin{tabular}{c|c|c}
    \toprule
      \textbf{Enforcement} & \textbf{Unsafe\%} & \textbf{Completion\%} \\  
      \midrule
      None  & 40.63\% & 59.38\% \\ \hline 
      \tool{}$^{0.9}_{stop}$ & 2.60\% & 10.42\% \\ \hline
      \tool{}$^{0.7}_{stop}$ & 5.20\% & 20.31\% \\ \hline
      \tool{}$^{0.5}_{stop}$ & 21.35\% & 41.14\% \\ \hline
      \tool{}$^{0.3}_{stop}$ & 29.17\% & 48.96\% \\ \hline
      \tool{}$^{0.9}_{reflect}$ & 14.07\%  & 47.74\% \\  
       
       \bottomrule
    \end{tabular}
    \label{tab:effectiveness_embodied}
\end{table}

\vspace{1mm}
\noindent\textbf{Effectiveness (RQ1):} 
We evaluate the performance of \tool{} in monitoring and enforcing safety properties within complex embodied environments. Table~\ref{tab:effectiveness_embodied} illustrates the inherent trade-off between safety violations (\textbf{Unsafe\%}) and functional utility (\textbf{Completion\%}) across various enforcement regimes. 

Without runtime monitoring, the agent exhibits a high violation rate of 40.63\%, while completing 59.38\% of tasks. \tool{} provides a configurable proactive defense through two primary intervention modes: \textit{stop} (immediate termination upon risk detection) and \textit{reflect} (risk-aware re-prompting). 
At the most conservative configuration (\tool{}$^{0.9}_{stop}$), the system successfully reduces unsafe outcomes to a negligible 2.60\%, a 93.60\% relative reduction over the unmonitored baseline. However, this comes at a cost: task completion drops to 10.42\% as the monitor prioritizes safety over progress. By scaling the threshold $\theta$, we observe a Pareto frontier of agent behavior: \tool{}$^{0.5}_{stop}$ provides a balanced middle ground, roughly doubling the completion rate of the next-strictest setting (41.14\% versus 20.31\% for \tool{}$^{0.7}_{stop}$) while maintaining violations at nearly half the baseline rate.  

Compared to the halt-based \textit{stop} strategy at the same threshold $\theta = 0.9$, \textit{reflect} allows the agent to maintain a much higher completion rate (47.74\%, i.e., 80.4\% of the unmonitored baseline, at a 65.37\% relative reduction in unsafe behavior) by attempting to self-correct its reasoning trace when a potential violation is predicted, shifting the agent from high-risk execution toward a safer ``self-correct'' posture.

\begin{figure}[t] 
        \centering
        \includegraphics[width=1\linewidth]{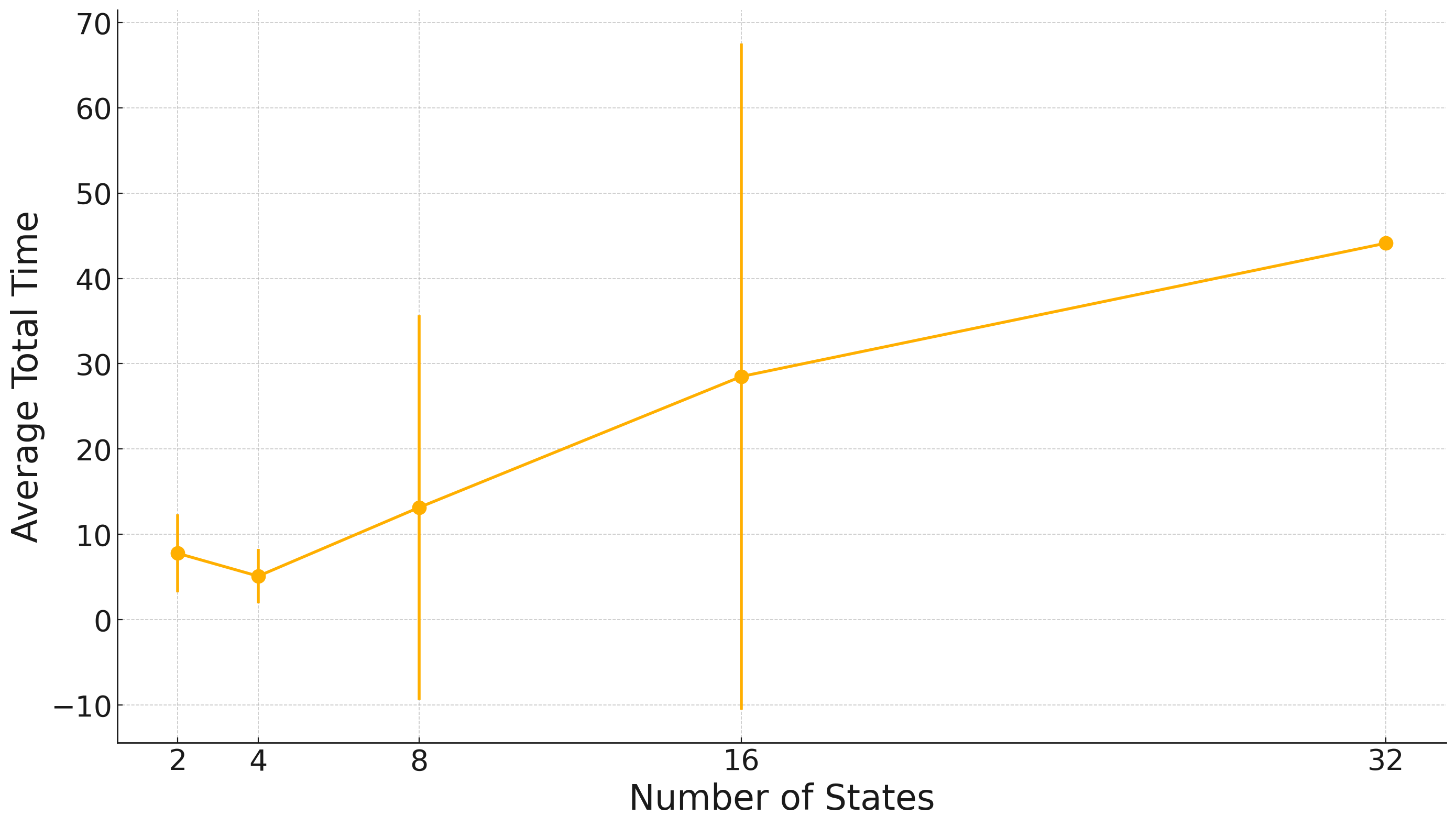}
        \caption{\rev{Average runtime overhead (milliseconds) with respect to the number of abstract states; bars indicate the variation observed across runs.}}
        \label{fig:overhead} 
\end{figure}

\smallskip
\noindent\textbf{Comparison with state of the art (RQ2):} 
We compare \tool{} with the state-of-the-art agent runtime enforcement framework AgentSpec~\cite{wang2025agentspec}. In addition to its effectiveness, the advantage of \tool{} is twofold.
(1) \emph{Runtime efficiency}: Unlike the reactive step-by-step enforcement in AgentSpec, which checks safety only after each LLM action, \tool{} performs predictive probabilistic reasoning over future paths.
By estimating the likelihood of a violation before the agent commits to an action, \tool{} early-rejects unsafe trajectories and avoids redundant LLM queries in long-horizon tasks, achieving an average token reduction of 12.05\%.
(2) \emph{Automated and trustworthy checks}: \tool{} automates the construction of CTL safety specifications systematically from unsafe-state definitions (e.g., ``stove on while agent away beyond threshold time''), eliminating the need for manually crafted rules.
In contrast, AgentSpec relies on manually engineered symbolic constraints that require substantial domain knowledge and per-task customization, incurring higher engineering overhead and limited scalability across new environments; its LLM-generated rules often lack interpretability and may be incomplete or incorrect.
The CTL formulas \tool{} synthesizes are instead grounded in a formally learned model of system behavior, and are transparent, verifiable, and admit probabilistic guarantees.

\vspace{1mm}
\noindent\textbf{Runtime overhead (RQ3):} 
We evaluate the runtime overhead of \tool{} by decomposing its enforcement process into abstraction, I/O, and probabilistic inference. The inference step (i.e., computing probability according to DTMC) is the dominant cost, averaging 430\,ms per decision cycle, while abstraction and I/O contribute only 0.07\,ms and 0.6\,ms, respectively. To reduce repeated inference costs, \tool{} employs a caching mechanism based on the fixed DTMC structure, precomputing the probability of reaching unsafe states for each symbolic state. This enables constant-time lookup during runtime, reducing the per-decision overhead to 5--8\,ms for small abstractions, 13\,ms for 8-state abstractions, and 28\,ms for 16-state abstractions, and remaining below 50\,ms at 32 states, as illustrated in Figure~\ref{fig:overhead}. The additional computation introduces only millisecond-level overhead, which is negligible relative to the LLM’s decision time.

\section{Discussion}  
\label{sec:discussion}

\smallskip
\rev{\noindent\textbf{PAC Bounds in Practice.}}
\label{sec:pac_discussion}
\rev{We analyze sample complexity following Bazille et al.~\cite{bazille2020global} with $\varepsilon{=}0.1$ and $\delta{=}0.05$. For the reachability property \tool{} checks at runtime, Theorem~3 of~\cite{bazille2020global} requires only $185$--$1,016$ traces via frequency estimation. For the stronger uniform CTL guarantee (Theorem~6), we measure the conditioning number $B(\hat{P}_\Pi)$ on the learned DTMCs (Table~\ref{tab:pac}). For the AV domain, although Laplace smoothing provides the stronger PAC guarantee by avoiding the assumption that unobserved transitions have zero probability, this requires an impractically large sample size.}

\begin{table}[t]
  \centering
  \caption{\rev{PAC bound analysis ($\varepsilon{=}0.1$, $\delta{=}0.05$). For AV, frequency estimation is used as a practical trade-off; the $B(\hat{P}_\Pi)$ column reports the conditioning number measured under Laplace smoothing, which is what makes the uniform bound impractical in that domain.}}
  \label{tab:pac}
  \begin{tabular}{lccc}
  \toprule
  Domain & $|S|$ & $B(\hat{P}_\Pi)$ & Traces required \\
  \midrule
  Embodied (with Laplace) & 3--33 & 1.0--481 & 530--$10^5$ \\
  AV (frequency estimation) & 4--34 & $10^{7}$--$10^{8}$ & 185--1{,}016 \\
  \bottomrule
  \end{tabular}
\end{table}

\rev{
The two domains exhibit a structural difference. Embodied-agent DTMCs model forward-progressing task execution (e.g., pick $\to$ open $\to$ place $\to$ close), producing well-mixed chains with moderate $B(\hat{P}_\Pi)$. In contrast, autonomous-driving DTMCs exhibit strong behavioral persistence: the vehicle remains within a narrow region of the abstract state space, with self-loop probabilities exceeding 99.9\% and inter-state transition probabilities below $10^{-5}$. Under Laplace smoothing, this persistence yields $B(\hat{P}_\Pi) \approx 10^7$--$10^8$, making the corresponding PAC bound impractical. By replacing Laplace smoothing with frequency estimation, the error factor introduced $\bigl(\tfrac{11}{10} B(\hat{P}_\Pi)\bigr)^2$ is removed from the bound, trading the stronger uniform CTL guarantee for a substantially lower sample requirement suitable for runtime monitoring.
}

\smallskip
\noindent\textbf{Threats to Validity.} A primary threat arises from the Markov assumption underlying our probabilistic model: the agent’s future behavior is assumed to depend only on the current abstract state rather than the full execution history. Although the predicate-based abstraction captures safety-relevant information, an incomplete or overly coarse abstraction may omit latent dependencies, such as delayed effects or history-dependent behavior. This can violate the Markov property, bias learned transition probabilities, and produce inaccurate risk estimates. Moreover, our PAC guarantees bound estimation error with respect to the learned model, but do not cover misspecification caused by an inadequate abstraction.

A second threat stems from stochasticity in trajectory sampling and agent behavior.  We mitigate this through repeated runs and report averages over those runs, although residual variance may still affect reproducibility and risk estimates, particularly in sparsely explored states. 

\rev{Our evaluation uses a fixed budget of 30 traces per scenario, below the bounds in Table~\ref{tab:pac}. The PAC analysis characterizes the sample complexity required for formal certification, whereas our experiments show that the monitor is practically effective with a much smaller budget. Closing this gap would require more extensive sampling and computational resources.}

\smallskip 
\noindent\textbf{Limitations and Future Work.}
Future work includes active learning and online updates for sample-efficient adaptation, improved predicate discovery and abstraction learning, support for richer PCTL properties, and natural language-to-specification pipelines. \rev{Scalability remains a key challenge as the predicate space grows; promising directions include abstraction refinement, compositional reasoning, symbolic model checking, and selective state-space exploration.}

\section{Related Work}
\label{sec:related_work}
\noindent\textbf{Agent Safety. }
This work contributes to the growing body of research on ensuring safe and reliable behavior in LLM-powered agents, a rapidly emerging field that combines runtime monitoring, and probabilistic reasoning to mitigate risks in open-ended decision-making. Recent benchmarks such as SafeAgentBench~\cite{yin2025safeagentbenchbenchmarksafetask}, AgentHarm~\cite{andriushchenko2024agentharmbenchmarkmeasuringharmfulness}, and AgentDOJO~\cite{debenedetti2024agentdojodynamicenvironmentevaluate} provide comprehensive testbeds for assessing agent behavior across diverse environments, including embodied tasks, simulated tool use, and interactive web settings.

To prevent agent risks, AgentSpec~\cite{wang2025agentspec} introduces a domain-specific language (DSL) for specifying symbolic runtime enforcement rules, enabling modular safety enforcement through structured prompt augmentations.
Other recent efforts, such as ShieldAgent~\cite{chen2025shieldagentshieldingagentsverifiable} and GuardAgent~\cite{xiang2025guardagentsafeguardllmagents}, propose shielding architectures that wrap LLM agents with logical constraints or policy filters. AgentDAM~\cite{zharmagambetov2025agentdamprivacyleakageevaluation} focus on privacy preservation in browser-based agents, highlighting the breadth of safety challenges faced by LLM systems. While effective for enforcing local invariants, these methods typically assume static constraints and ignore multi-step risk evolution. In contrast, our approach combines symbolic abstraction with probabilistic reachability for trajectory-aware enforcement, extending safe RL shielding~\cite{alshiekh2018safe} to data-driven, stochastic LLM agents through learned probabilistic runtime monitoring.

Complementary research has also explored constraining LLM behavior through formal logic and decoding-time control. For example, LMQL~\cite{beurer2023lmql} introduces a query language that enforces logical constraints during decoding, ensuring syntactic and semantic compliance at generation time. However, LMQL’s guarantees are output-level and static, whereas our approach operates dynamically at runtime, reasoning over symbolic state transitions and enforcing safety through probabilistic reachability analysis. Similarly, frameworks such as Toolformer~\cite{schick2023toolformer} and Voyager~\cite{xu2023voyager} have expanded the scope of LLM agents into tool-augmented and open-world environments, amplifying the need for proactive safety enforcement that anticipates multi-step, high-risk behaviors before they manifest.

\vspace{2mm}
\noindent\textbf{Runtime Verification}
Our work builds on the runtime verification (RV) literature while extending it to provide safety and reliability guarantees for LLM-powered agents operating in uncertain environments. Classical RV monitors system executions against formal specifications and triggers interventions upon detecting violations~\cite{leucker2009brief,bartocci2018introduction}. However, traditional RV typically assumes deterministic and fully observable system dynamics, limiting its applicability to stochastic agent workflows. 
To address these limitations, several probabilistic extensions have been proposed. Runtime verification with state estimation (RVSE)~\cite{stoller2011rvstate} augments monitoring with probabilistic inference over hidden states, while adaptive RV~\cite{bartocci2012arv} updates monitoring models online to accommodate evolving system behaviors. \rev{Another related direction combines conformal prediction with runtime verification to forecast future states from observed executions while providing distribution-free coverage guarantees~\cite{lindemann2023conformal,cairoli2023learning}.} Similarly, probabilistic monitoring frameworks such as PSTMonitor~\cite{burlò2022pstmonitor} and MDP-based monitors~\cite{junges2021mdpmonitor} incorporate stochastic reasoning into runtime monitoring, but they assume either predefined probabilistic models or manually specified safety properties.
Our framework differs by learning the probabilistic model directly from observed agent trajectories. Specifically, we construct a DTMC over predicate-based symbolic states that capture high-level safety conditions, enabling data-driven and domain-general runtime enforcement without requiring a handcrafted transition model. During execution, probabilistic reachability analysis estimates the likelihood of future safety violations, allowing proactive intervention before unsafe states are reached.
\section{Conclusion}
\label{sec:conclusion}
We presented \tool{}, a proactive runtime enforcement framework that enhances
LLM-agent safety through probabilistic prediction. By modeling agent behavior
as DTMCs over symbolic abstractions, \tool{} anticipates risks and can intervene
before violations occur. Experiments in embodied-agent and autonomous-driving
domains show that \tool{} can improve safety while preserving task performance,
offering a principled approach that is adaptable across dynamic agent environments.

\section*{Acknowledgments}
This research is supported by the Lee Kuan Yew Fellowship awarded to SUN Jun by Singapore Management University. 
We sincerely thank the anonymous reviewers for their valuable feedback and suggestions.

\section*{Data Availability Statement}
The artifacts generated and used in this study---including the benchmark,
safety rules, learned DTMCs, and source code---are available in the repository
at~\cite{wang2026probguard_code}. The repository also includes our
LangChain-based implementation of the runtime monitoring and intervention
mechanisms to support reproduction.

\bibliographystyle{ACM-Reference-Format}

\bibliography{citations}

\appendix
\section{Proof of Theorem~\ref{thm:pac}}
\label{app:proof}

We restate the theorem before giving the proof. Throughout, Theorems~5 and~6
refer to the results of Bazille et al.~\cite{bazille2020global}, not to results
of this paper.

\medskip
\noindent\textbf{Theorem~\ref{thm:pac}.}
\emph{For any safety property $\psi=\mathbf{A}\,\psi^{\pi}$ whose path formula
$\psi^{\pi}$ is a reachability or invariant property of the form covered
by~\cite{bazille2020global}, confidence $\delta$, and error $\varepsilon$,
Algorithm~\ref{alg:learn_matrix_ctl} is $(\varepsilon,\delta)$-PAC-correct.}

\begin{proof}
Let $\mathcal{M}=(S,P)$ be the (unknown) DTMC governing the agent's behavior, and
let $\hat{\mathcal{M}}=(S,\hat P_\Pi^\alpha)$ be the DTMC returned by
Algorithm~\ref{alg:learn_matrix_ctl}.
Fix a property $\psi=\mathbf{A}\,\psi^{\pi}$ as in the statement, accuracy~$\varepsilon$,
and confidence~$\delta$. Set
\[
\delta'=\frac{\delta}{|S|},
\qquad
\varepsilon'=\frac{\varepsilon}{B(\hat P_\Pi^\alpha)} .
\]

\paragraph{Step~1: Transition-probability concentration.}
For each state $s\in S$, the algorithm maintains the pairwise transition counts
$n_{st}^\Pi$ and the visitation count $n_s^\Pi=\sum_{t} n_{st}^\Pi$.
Define the state-wise threshold
\[
H^\ast(n_s^\Pi,\varepsilon',\delta')
    =
    \max_{t\in S}
    H\bigl(n_s^\Pi, n_{st}^\Pi, \varepsilon', \delta'\bigr),
\]
where $H(\cdot)$ is the transition-wise concentration bound of Theorem~6
of~\cite{bazille2020global}. By the stopping rule of
Algorithm~\ref{alg:learn_matrix_ctl}, when the algorithm terminates,
\[
n_s^\Pi
\;\ge\;
\Bigl(\tfrac{11}{10} B(\hat P_\Pi^\alpha)\Bigr)^2
H^\ast(n_s^\Pi,\varepsilon',\delta'),
\qquad \forall s\in S.
\]
Thus, by Theorem~6 of~\cite{bazille2020global}, for each fixed $s$, with
probability at least $1-\delta'$,
\[
\max_{t\in S}
\bigl|P(s,t)-\hat P_\Pi^\alpha(s,t)\bigr|
\;\le\;
\varepsilon' .
\]
Applying a union bound over the $|S|$ source states, and using
$\delta'=\delta/|S|$, gives
\begin{equation}
\Pr\!\left(
    \max_{s,t\in S}
    \bigl|P(s,t)-\hat P_\Pi^\alpha(s,t)\bigr|
    \le \varepsilon'
\right)
\ge 1-\delta .
\label{eq:good-transition-event}
\end{equation}

\paragraph{Step~2: Propagating local error to satisfaction probabilities.}
By Theorem~5 of~\cite{bazille2020global} (reachability sensitivity), for any path
property $\psi^{\pi}$ in the fragment covered by that result,
\[
\bigl|
    \mathbb{P}_{\mathcal{M}}(\psi^{\pi})
    -
    \mathbb{P}_{\hat{\mathcal{M}}}(\psi^{\pi})
\bigr|
\;\le\;
B(\hat P_\Pi^\alpha)\cdot
\max_{s,t\in S}\bigl|P(s,t)-\hat P_\Pi^\alpha(s,t)\bigr|.
\]
The conditioning number is evaluated on the learned chain
$\hat P_\Pi^\alpha$ rather than on $P$; this is what makes the bound
checkable at runtime, and is admissible because Laplace smoothing keeps every
semantically valid transition bounded away from zero, so
$B(\hat P_\Pi^\alpha)$ is finite and computable from the data the algorithm
already maintains.

Under the good event of Eq.~\eqref{eq:good-transition-event},
\[
\max_{s,t}|P(s,t)-\hat P_\Pi^\alpha(s,t)|
    \;\le\; \varepsilon'
    = \frac{\varepsilon}{B(\hat P_\Pi^\alpha)},
\]
implying
\[
\bigl|
    \mathbb{P}_{\mathcal{M}}(\psi^{\pi})
    -
    \mathbb{P}_{\hat{\mathcal{M}}}(\psi^{\pi})
\bigr|
\;\le\;
B(\hat P_\Pi^\alpha)\,\varepsilon'
\;=\;
\varepsilon.
\]

\paragraph{Step~3: PAC guarantee.}
Combining the above yields
\[
\Pr\!\left(
\left|
    \mathbb{P}_{\hat{\mathcal{M}}}(\psi^{\pi})
    -
    \mathbb{P}_{\mathcal{M}}(\psi^{\pi})
\right|
\le \varepsilon
\right)
\;\ge\;
1-\delta,
\]
which is precisely $(\varepsilon,\delta)$-PAC correctness
(Definition~\ref{def:pac}). Thus
Algorithm~\ref{alg:learn_matrix_ctl} is PAC-correct.
\end{proof}

\end{document}